\documentclass[lettersize,journal]{IEEEtran}
\usepackage{amsmath,amsfonts}
\usepackage{algorithmic}
\usepackage{algorithm}
\usepackage{array}
\usepackage[caption=false,font=normalsize,labelfont=sf,textfont=sf]{subfig}
\usepackage{textcomp}
\usepackage{stfloats}
\usepackage{url}
\usepackage{verbatim}
\usepackage{graphicx}
\usepackage{cite}
\usepackage[hidelinks]{hyperref}
\usepackage{svg}

\begin{document}

\title{EXPRTS: Exploring and Probing the Robustness of Time Series Forecasting Models}
\author{Håkon Hanisch Kjærnli,
\and Lluis Mas-Ribas,
\and Hans Jakob Håland,
\and Vegard Sjåvik,
\and Aida Ashrafi,
\and Helge Langseth,
\and and Odd Erik Gundersen

\thanks{H. H. Kjærnli is with Aneo AS and the Norwegian University of Science and Technology. Email: hakon.hanisch.kjernli@aneo.com.}
\thanks{L. Mas-Ribas is with the University of California Santa Cruz. Work done while at the Norwegian University of Science and Technology and Aneo AS.}
\thanks{H. J. Håland is with the Norwegian University of Science and Technology.}
\thanks{V. Sjåvik is with the Norwegian University of Science and Technology.}
\thanks{A. Ashrafi is with the University of Bergen. Work done while at Aneo AS.}
\thanks{H. Langseth is with the Norwegian University of Science and Technology and Aneo AS.}
\thanks{O. E. Gundersen is with the Norwegian University of Science and Technology and Aneo As.}}

\markboth{Kjærnli \MakeLowercase{\textit{et al.}}: EXPRTS: Exploring and Probing the Robustness of Time Series Forecasting Models}
{Kjærnli \MakeLowercase{\textit{et al.}}: EXPRTS: Exploring and Probing the Robustness of Time Series Forecasting Models}

\maketitle

\begin{abstract}
When deploying time series forecasting models based on machine learning to real world settings, one often encounter situations where the data distribution drifts. Such drifts expose the forecasting models to out-of-distribution (OOD) data, and machine learning models lack robustness in these settings. Robustness can be improved by using deep generative models or genetic algorithms to augment time series datasets, but these approaches lack interpretability and are computationally expensive. In this work, we develop an interpretable and simple framework for generating time series. Our method combines time-series decompositions with analytic functions, and is able to generate time series with characteristics matching both in- and out-of-distribution data. This approach allows users to generate new time series in an interpretable fashion, which can be used to augment the dataset and improve forecasting robustness. We demonstrate our framework through EXPRTS, a visual analytics tool designed for univariate time series forecasting models and datasets. Different visualizations of the data distribution, forecasting errors and single time series instances enable users to explore time series datasets, apply transformations, and evaluate forecasting model robustness across diverse scenarios. We show how our framework can generate meaningful OOD time series that improve model robustness, and we validate EXPRTS effectiveness and usability through three use-cases and a user study.
\end{abstract}


\section{Introduction}

\begin{figure*}[ht]
    \centering
    \includegraphics[width=\textwidth]{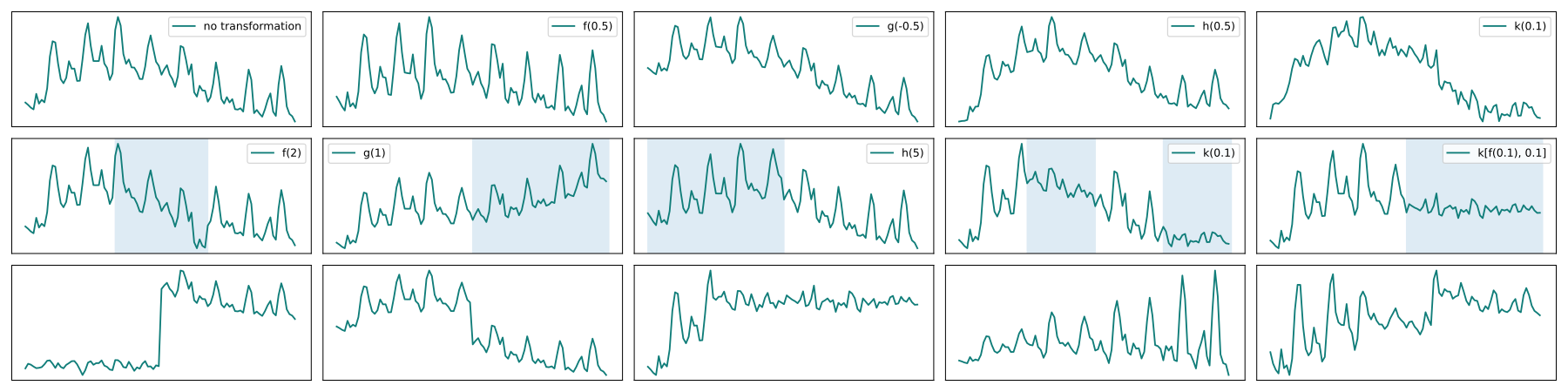}
    \caption{Examples of different time series generated with our framework. The first row shows transformations possible using the work of Kegel et al.\cite{kegel_generating_2017}, with the reference time series shown in the top left corner. The second row shows transformations applied to the same reference time series, but only transforming observations within the highlighted interval. In the last row we use our framework to transform the reference time series with several different types of transformations applied at multiple intervals.}
    \label{fig:transformations}
\end{figure*}

When machine learning models are deployed in real-world settings, they may face a challenge in the form of an evolving data distribution: over time, the distribution of the input data or the relationship between input and output change\cite{quionero-candela_dataset_2009, moreno-torres_unifying_2012}. Data that originates from a distribution different than the distribution of the training dataset is referred to as out-of-distribution data, and machine learning models are known to lack robustness when applied to such data; model performance degrades and the output cannot be trusted \cite{42503, hendrycks_benchmarking_2018, belinkov_synthetic_2018}.

Sometimes it is possible to predict, or even know, how the data distribution will change. Real-life examples include shifts in patient demographics in healthcare\cite{sahiner_data_2023}, or increased electricity price volatility as more variable renewable energy is integrated into the power grid\cite{impram_challenges_2020}. In these situations, models can be made more robust by augmenting the training dataset with data that reflect the anticipated changes\cite{shankar2018generalizing, shrivastava_learning_2017}. However, such an approach requires a procedure to generate data with characteristics and semantics that match those we expect to encounter in the future\cite{hoffman_cycada_2018}.

Here, our focus is on the task of univariate time series forecasting and we develop a framework that allows us to generate time series with a wide variety of characteristics. Deep generative models like GANs\cite{goodfellow_generative_2014}, VAEs\cite{kingma_auto-encoding_2014} and Diffusion models\cite{sohl-dickstein_deep_2015} have been shown to be effective at generating new and realistic data. Unfortunately, these models are inherently unsuitable for generating out-of-distribution data as they are designed to model data from the training distribution. Furthermore, they are based on large neural networks and trained with complex procedures making it hard for users to explain and interpret how data is generated. An alternative approach uses genetic algorithms to evolve a pool of time series such that they exhibit some targeted set of characteristics\cite{kang_visualising_2017, kang_gratis_2020}. While genetic algorithms can produce out-of-distribution data with specific characteristics and semantics, the data generating procedure is not interpretable and it remains complex and computationally expensive.

In contrast to deep generative models and genetic algorithms, we base our method on simple, interpretable transformations that are easy for users to understand. Specifically, we generate time series data by first decomposing an existing series, separating it into independent components with a well known semantic meaning. Then, building on \cite{kegel_generating_2017}, our framework allows users to transform the components independently and at arbitrary time intervals. The transformation are designed to change specific characteristics of a time series, giving users control of the kind of data they generate. As seen in Figure \ref{fig:transformations}, the end result is a framework that can generate a diverse set time series.

To showcase our framework we build \texttt{EXPRTS}, a tool that lets users visualize time series datasets and probe the robustness of forecasting models. Frameworks and tools built to enable users to explore, explain and interpret datasets or machine learning models are studied in the field of Visual Analytics (VA)\cite{hohman_visual_2019}. In the context of machine learning, VA tools combine techniques from eXplainable Artificial Intelligence (XAI) and visualization to let users identify patterns or issues within a dataset, or gain insights into the inner workings of a model\cite{cheng_dece_2021, kaul_improving_2022, yeh_attentionviz_2024}. By involving users into the development and evaluation process, such tools can help users find and correct issues in a dataset or model\cite{8807255}. As a result, models can be iteratively refined to become more robust in key situations\cite{spinner_explainer_2020}. \texttt{EXPRTS} enables a similar kind of workflow: users can explore datasets, check model performance with different error metrics and at different forecasting horizons, and our transformations allow users to probe and improve model robustness.

In \texttt{EXPRTS}, we describe each time series in a dataset using four features. These four features are then embedded into a two-dimensional space using principal component analysis (PCA), creating an instance space\cite{kang_visualising_2017}. Here, we construct the instance space from features that correspond directly with the characteristics that our framework is designed to transform, producing a strong semantic link between the instance space and the transformations applied by the user. By combining the instance space and the transformations, \texttt{EXPRTS} lets users select and transform individual time series, explore the instance space and evaluate model performance on data with different characteristics. To demonstrate the usefulness of our framework, we show that it is able to generate out-of-distribution data that makes a forecasting model more robust. Lastly, we present the functionality of \texttt{EXPRTS} through three use-cases, and evaluate the tool through user testing.

To summarize, our contributions are:
\begin{itemize}
    \item A fast and simple framework for generating time series data.
    \item We present \texttt{EXPRTS}, a tool that allows users to explore time series datasets and probe a forecasting model's robustness.
    \item We conduct a user study, validating the usefulness of our tool and the simple, interpretable methods used to visualize the datasets and transform time series. The results show that our simple approach is indeed useful, and that it provides insights about datasets and models, and enable users to probe the robustness of machine learning models applied to time series forecasting.
\end{itemize}

\section{Related Work}
Numerous visual analytics tools have been developed to allow users to explore and explain machine learning models and datasets. The goal of such tools is often explain and interpret a model. By doing so, the tools can enhance users' understanding of the model, help guide further model development or provide decision support. Visual analytics tools that investigate model performance when exposed to different inputs include tools like DECE\cite{cheng_dece_2021}; a tool using counterfactuals\cite{wachter_counterfactual_2018} to visualize model prediction on both individual instances and data subsets. Another example is CoFact\cite{kaul_improving_2022}, which aims at investigating causality and the relationship between features. Explainer\cite{spinner_explainer_2020} is similar to the two aforementioned tools, but focuses on an iterative approach that allow users to gradually interpret and refine machine learning models. Of tools not specific for time series, the What-If Tool\cite{8807255} is perhaps the one most closely resembling ours. The tool is model and dataset agnostic, and allows users to explore and probe both datasets and model outputs. However, contrary to ours, it is not designed to handle and visualize time series data.

VA tools for time series must be able to visualize sequences, possibly spanning both space and time. Considering temporal data only, \cite{ali_timecluster_2019, rodriguez-fernandez_deepvats_2023} present tools that lets users explore datasets by using an autoencoder for dimensionality reduction. The MultiRNNExplorer\cite{shen_visual_2020} also use an autoencoder to learn a latent space that can be visualized, but with a focus on visualizing the representations learned by a neural network instead of the dataset. Spatio-temporal data is handled in \cite{li_cope_2019, fujiwara_visual_2021}, with both works utilizing dimensionality reduction to let users identify abnormal subsequences in a dataset.

If we consider tools specifically made for time series forecasting, many examples revolve around assisting users in the model selection process. The work in \cite{bogl_visual_2013} presents a tool targeted towards traditionally time series analysis and models, providing  Q-Q plots and information about autocorrelation and partial autocorrelation. In \cite{sun_dfseer_2020}, the authors present a tool for assisting users in the model selection process by allowing models to be ranked based on different metrics and display performance on various subsets of data. \cite{xu_mtseer_2021} also focuses on model selection, but includes techniques from XAI to explain model behavior, adding another dimension to the model selection process. The tools in \cite{10.1162/neco.1997.9.8.1735, 10297593} are more directly related to our work and both uses counterfactuals in combinations with a LSTM\cite{10.1162/neco.1997.9.8.1735}. However, the tool in \cite{lee_visual_2020} is specifically designed for traffic congestion and while TimeTuner\cite{10297593} is more general, it differs from our work in the sense that the objects of interest are the internal representations of a neural network.

Most of the aforementioned related work concerns finding a representation for time series which are then visualized, allowing users to identify clusters of similar instances. As evident by the plethora of suggested methods to solve the task, finding good representations is difficult. Instead of focusing on finding these representations, we take a simpler approach and use time series instance spaces\cite{kang_visualising_2017}. Instance spaces have been used in several interesting applications and analysis within the field of time series forecasting\cite{kang_gratis_2020, spiliotis_are_2020, talagala_meta-learning_2023, talagala_fformpp_2022, semenoglou_data_2023}. The method condenses a time series into a set of characteristics with known, interpretable semantics, making the resulting representation easy to understand. Additionally, the characteristics are usually independent of the length of the time series and thus suitable for visualizing instances of different lengths. The simplicity of the instance spaces makes our tool flexible and allows us explore most kinds of models and datasets used to benchmark forecasting methods.

\section{Time-series Features and Transformations}\label{sec:ts-features}
Let $\{x_i\}^T_{i=1}$ be a sequence of observations of a time series, and denote $i$th observation of the series as $x_i$. A key assumption in our work is that $x_i$ can be  decomposed into a trend component, a seasonal component and a remainder component. A time series contains a trend component if the level of the series changes time, while a seasonal component is present if the time series is affected by factors like the time of year or day of week. The remainder is simply any remaining values left after subtracting the estimated trend and seasonal component from the time series. Here we use STL \cite{cleveland90} to decompose time series. Letting $t_i$, $s_i$ and $r_i$ denote the trend, seasonal and remainder components of the $i$th observation, we decompose $x_i$ as $x_i = t_i + s_i + r_i$.

Datasets used for research on machine learning for time series often contain a large number of related time series. We represent these distinct time series using a set of features, allowing us use dimensionality reduction techniques to create an instance space that can be used to visualize the dataset. Finding a suitable set of features is important to create an informative instance space. For the dataset we consider, we found the features used in \cite{kegel_generating_2017} useful, and these features have, in combination with others, been used to create instance spaces for similar datasets\cite{kang_visualising_2017, kang_gratis_2020, semenoglou_data_2023, spiliotis_are_2020}. The four features represent different aspects of the decomposed time series: the trend strength, the seasonal strength, trend linearity and trend slope.

The trend strength and seasonal strength are defined using similar equations. First, let $\bar{x} = \frac{1}{T}\sum^T_{i=1} x_i$ denote the sample mean of a sequence of observations. The sample variance is then denoted using $\text{Var}(x) = \frac{1}{T - 1}\sum^T_{i=1} (x_i - \bar{x})^2$. Letting $F_1$ denote the trend strength we define:

\begin{equation}\label{eq:t-str}
    F_1 = \max\left(0, 1 - \frac{\text{Var}(r)}{\text{Var}(t + r)}\right),
\end{equation}

\noindent
and define the seasonal strength $F_2$ as:

\begin{equation}\label{eq:s-str}
    F_2 = \max\left(0, 1 - \frac{\text{Var}(r)}{\text{Var}(s + r)}\right).
\end{equation}

\noindent
$F_1$ and $F_2$ are similar, but $F_1$ compares the sample variance of the remainder component with the deseasonalized series, $\{t_i + r_i\}^T_{i=1}$. $F_2$ on the other hand, compares the sample variance of the remainder with the detrended series, $\{s_i + r_i\}^T_{i=1}$. Both $F_1$ and $F_2$ are numbers between 1 and 0, where 1 represents a time series with a strongly pronounced trend or seasonal component.

To define the last two features, we express $i$th value of the trend component through a linear regression model. Let $\beta_0$ and $\beta_1$ denote the coefficients in a linear regression model fit to $\{t_i\}^T_{i=1}$, and $\delta_i$ the deviation from $t_i$ to the fitted line. Then we can write:

\begin{equation}\label{eq:lin-reg}
    t_i = \beta_0 + \beta_1 \cdot i + \delta_i.
\end{equation}

\noindent
Equation \ref{eq:lin-reg} enables us to characterize the trend linearity and slope. Letting $\text{Var}(\delta)$ denote the sample variance of the sequence of deviances $\{\delta_i\}^T_{i=1}$, the trend linearity feature is given by:

\begin{equation}\label{eq:t-lin}
    F_3 = \max\left(0, 1 - \frac{\text{Var}(\delta)}{\text{Var}(t)}\right).
\end{equation}

\noindent
$F_3$ is close to 1 when the sample variance of the deviances $\{\delta_i\}^T_{i=1}$ is low compared to the sample variance of the trend component $\{t_i\}^T_{i=1}$.

The trend slope $F_4$ is simply defined as the slope of the regression model in Equation $\ref{eq:lin-reg}$:

\begin{equation}\label{eq:t-slope}
    F_4 = \beta_1.
\end{equation}

\begin{figure*}[t]
    \centering
    \includegraphics[width=\textwidth]{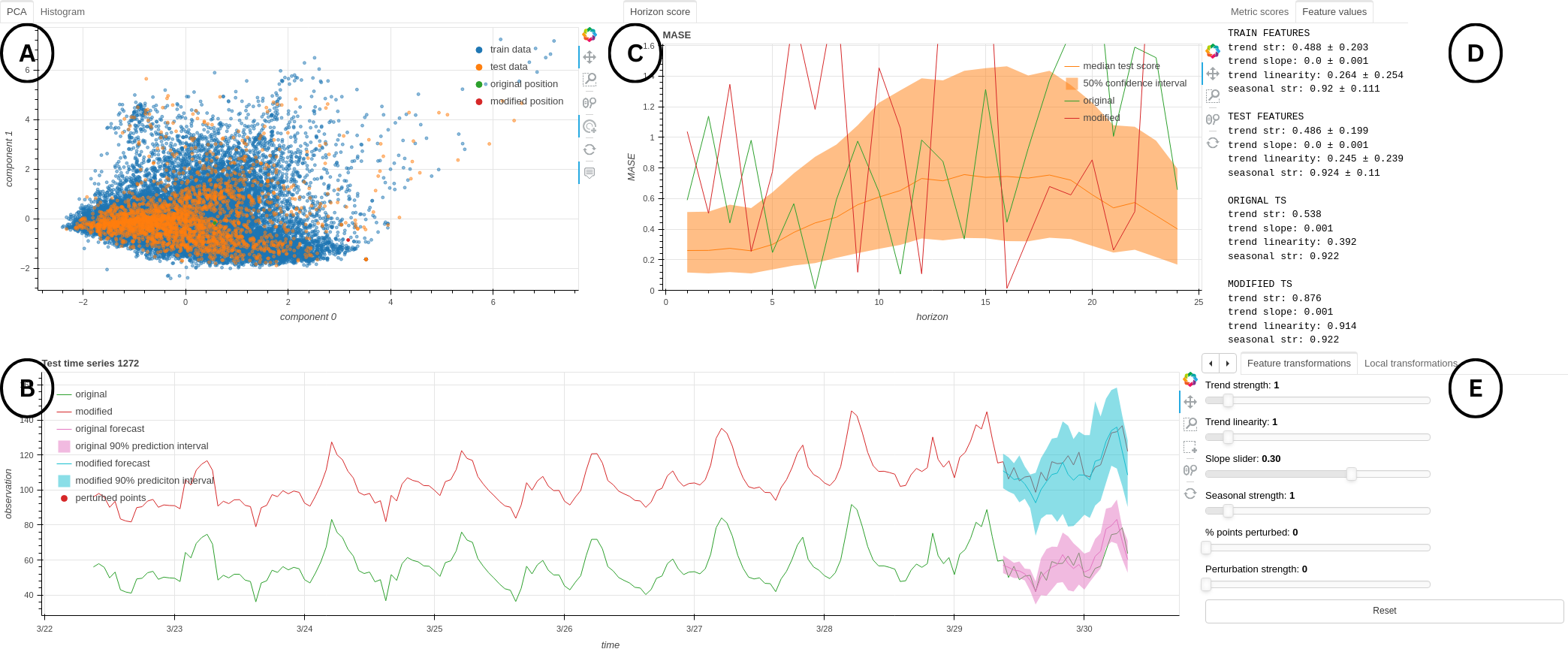}
    \caption{The interface of \texttt{EXPRTS} when applied to the electricity dataset. Instances of the dataset is visualized in (A). Users choose between a PCA-plot (shown in the figure) or a histogram representation through the tab above (A). Selecting a single time series makes it appear in (B), which, together with (A), provides users with an overview of both the dataset and individual instances. Panel (C) lets users inspect the errors of the forecasting model at different forecasting horizons. Panel (D) allow users to inspect numeric values related to the selected time series. Users can change between the numeric values of the feature values (as shown in the plot) or error metrics by switching the tab above (D). The selected time series can be transformed by using the different sliders and options available in (E), prompting changes in the other panels. Multiple transformation options are found in the other tabs of (E), as well as other UI-elements that allow users to select different error metrics and specific time series.}
    \label{fig:ui}
\end{figure*}

\noindent
Calculating $F_1$, $F_2$, $F_3$, and $F_4$ allows us to create a compact and interpretable representation for time series. We calculate these features for all time series in a dataset, and then use PCA to project the dataset down to two dimensions. Figure \ref{fig:ui} (panel A) shows an example of the instance space that the procedure creates. Instance spaces have been used to identify outliers in time series datasets\cite{kandanaarachchi_normalization_2020, talagala_feature-based_2019}, to generate time series data and analyze the performance of different forecasting models\cite{kang_visualising_2017, kang_gratis_2020}, and Semenoglou et al.\cite{semenoglou_data_2023} used instance spaces to investigate different techniques for time series data augmentation.

To generate time series data, we build on the transformations defined in \cite{kegel_generating_2017}. The first transformation modifies the trend strength using three parameters: $f$, $h$, and $m$. Let $\tilde{t}_i$ denote the transformed trend component for observation $i$. By expressing the trend component through the linear regression model in Equation \ref{eq:lin-reg}, we can transform the $i$th element of the component using the following equation:

\begin{equation}\label{eq:t-transform}
    \tilde{t}_i = \beta_0 + f \cdot \left(\beta_1 \cdot i + \frac{\delta_i}{h} \right) + \; m \cdot \beta_0 \cdot i.
\end{equation}

\noindent
In Equation \ref{eq:t-transform}, $f$ changes the trend strength $F_1$. For $f > 1$, $F_1$ increases and for $f < 1$, $F_1$ decreases. The parameter $h$ works similarly, but changes the trend linearity $F_3$. Large values of $h$ will increase the trend linearity by decreasing the deviation away from the fit of the linear model. With parameters $f$ and $h$ we are able to modify the trend strength and the trend linearity, and changing the slope feature $F_4$ could be done by simply changing the coefficient $\beta_1$. However, note that in the case of very small trend component, doing so would require a very large parameter value. Instead we do as proposed in \cite{kegel_generating_2017} and transform the trend slope by multiplying the intercept with a parameter $m$, which is then added to the final transformed trend component. A positive slope is added to the time series if $m > 0$, while $m < 0$ adds a negative slope. 

Finally, we let $\tilde{s}_i$ denote the transformed seasonal component for the $i$th observation. We transform the seasonal component with the following equation:

\begin{equation}\label{eq:s-transform}
    \tilde{s}_k = k\cdot s_k.
\end{equation}

\noindent
Equation \ref{eq:s-transform} changes the seasonal strength $F_2$ through the parameter $k$, effectively changing the amplitude of the seasonal component.

After transforming the trend and seasonal components, the new time series is created by reassembling it from the individual components. Letting $\tilde{x}_i$ denote the transformed time series, we have $\tilde{x}_i = \tilde{t}_i + \tilde{s}_i + r_i$.  As seen in the top row of Figure \ref{fig:transformations}, the transformations are able to change time series in a wide variety of different ways, and their simplicity ensure that they are easy for end users to understand and interpret. However, they only apply to full-length time series and individual components, limiting the kind of data that can be generated. For example, it is not possible to generate time series that exhibit sudden jumps in level or changes in the seasonal pattern. We address these limitations in two ways: 1) we allow transformations to be applied locally and at arbitrary indexes; and 2) by introducing two simple, but fundamental, transformations. Specifically, we include translation of the mean level and perturbations with Gaussian noise as transformations, and we enable these, and the four other transformations, to be applied to subsets of a time series. Doing so gives us the ability to generate a large set of interesting time series, with row 2 and 3 of Figure \ref{fig:transformations} showing some examples. Despite the increased generality of the framework, generating new time series remains fast and simple.

\section{EXPRTS}\label{sec:tool}

Here we present the \texttt{EXPRTS} tool, which we make publicly available to the community at \url{https://github.com/AneoGroup/EXPRTS}. First, in Section \ref{sec:goals} we provide an overview of the underlying design requirements that lead to the development of \texttt{EXPRTS}. Section \ref{sec:sys-overview} describes the system architecture and workflow. Lastly, in Section \ref{sec:gui} we provide an explanation of the application and each individual UI element.

\subsection{Design Requirements}\label{sec:goals}
\texttt{EXPRTS} was developed with expert users in mind, and with a explicit goal of enabling exploration and analysis of machine learning models applied to time series forecasting. The design requirements are motivated by the target users and goals of the tool, and, as discussed in \cite{gneiting_making_2011}, the difficulty of evaluating time series forecasts. Thus, our aim is to:

\begin{description}
    \item[\textbf{R1}] \textbf{Visualize and explore time series datasets.} The tool must be able to visualize both single time series and entire datasets. The visualization should help users understand the characteristics present in a given dataset and how the individual time series relate to each other.
    \item[\textbf{R2}] \textbf{Explore model performance.} Assessing model performance is hard and is, in part, the motivation of time series instance spaces\cite{kang_visualising_2017}. To assist users in this task, the tool must visualize model performance in multiple ways, both quantitative and qualitative, and through multiple metrics.
    \item[\textbf{R3}] \textbf{Explore model robustness.} Understanding the types of time series a model is able to reliably forecast is critical in real-life application of time series forecasting. \texttt{EXPRTS} should give users insights about model robustness by allowing time series to be transformed, enabling assessment of robustness in specific scenarios.
\end{description}

\subsection{System overview}\label{sec:sys-overview}

\begin{figure}[t]
    \centering
    \includegraphics[width=\linewidth]{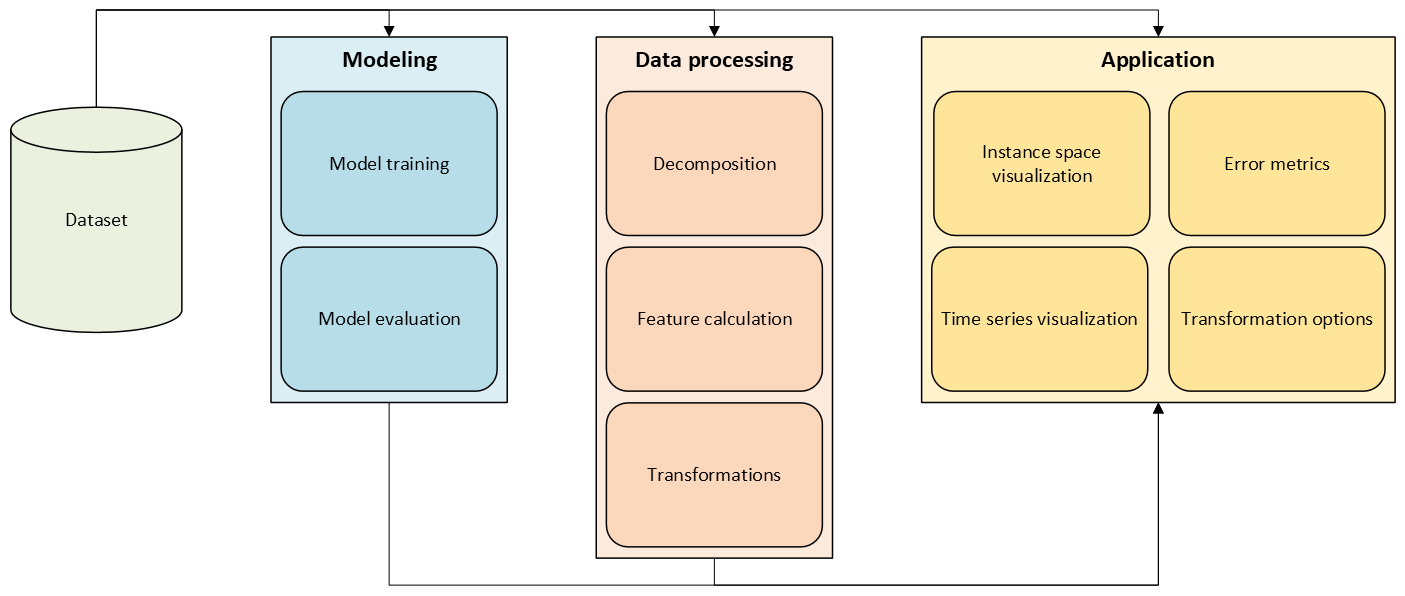}
    \caption{An overview of the modules of \texttt{EXPRTS}. Users select a datasets, which is then used as input for the individual modules. The dataset, together with the the outputs of the modeling and data processing modules, is used by the application module.}
    \label{fig:architecture}
\end{figure}

An overview of the software architecture of \texttt{EXPRTS} can be seen in Figure \ref{fig:architecture} and is divided into three main modules: the modeling module, the data processing module and the application module. The application module depends on the output of the modeling and data processing modules as well as the raw dataset; the modeling and data processing modules depend only on the dataset. To run \texttt{EXPRTS}, users first select a univariate time series dataset. The dataset can contain one single time series or multiple. For convenience, we have integrated our application tightly with the datasets available through GluonTS\cite{gluonts_jmlr}, but our tool also supports using local datasets.

After choosing a dataset, the user must select a model. \texttt{EXPRTS} comes with four models already implemented: a simple dense neural network, N-BEATS \cite{oreshkin_n-beats_2020}, a temporal convolutional network (TCN) \cite{oord_wavenet_2016} in the form of the MQ-CNN forecaster by Wen et al. \cite{wen_multi-horizon_2018}, and a basic Transformer based on\cite{vaswani_attention_2017}. All of them implemented by making use of Pytorch\cite{paszke_pytorch_2019}. The tool is designed to be model agnostic and easily extensible to other models; we have defined a simple interface similar to the model interfaces found in Scikit-Learn\cite{pedregosa_scikit-learn_2011}. Defining such an interface allows users to import their own models into \texttt{EXPRTS} with minimal work while retaining flexibility. Once the user has chosen a model, the model is trained and evaluated to enable visualizations of errors and predictions in the user interface.

The data processing module contains three submodules responsible for the decomposition of the time series, the calculation of the features and the transformation of time series. Having selected a dataset, the application requires the individual time series instances to be described using a set of features. To do so, each instance is first decomposed before the feature values of the instance is calculated. The transformations submodule contains the functionality for transforming time series. All of the submodules of the data processing module work as described in Section \ref{sec:ts-features}.

In the application module we find the components of our tool related to the user interface. At a high level, the module contains a submodule to visualize the instance space resulting from the feature calculation in the data procesing module; a submodule to represent the error metrics calculated in the modeling module; a time series visualization submodule to provide a visual plot of a specific time series of interest; and lastly, a submodule that allows the user to apply transformations to a selected time series.

\subsection{The User Interface} \label{sec:gui}
\begin{figure}
    \centering
    \includegraphics[width=1\linewidth]{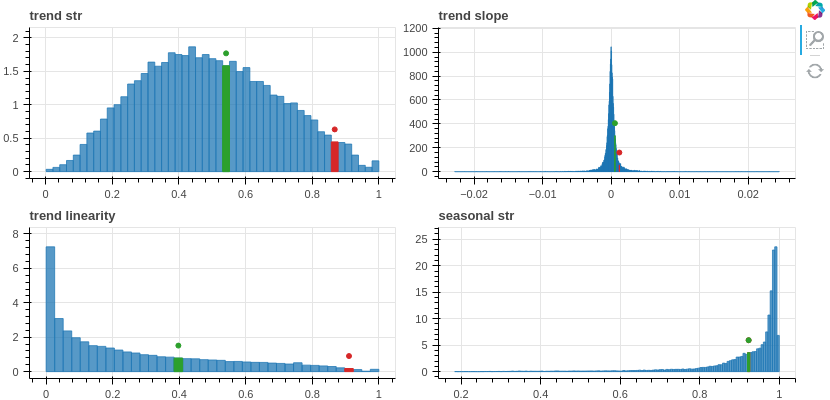}
    \caption{The histogram view of the dataset. Shown if users select the histogram tab in Figure \ref{fig:ui}, panel (A). The bin of the original time series is colored green, and the bin of the transformed time series is colored red.}
    \label{fig:histogram}
\end{figure}

Figure \ref{fig:ui} shows the \texttt{EXPRTS} graphical user interface (GUI) built with the visualization tool Bokeh\cite{bokeh_development_team_bokeh_2018}. Through the GUI, the user can visualize the data distribution, select and display individual time series, transform their features, and inspect the prediction scores for time series.

The top left panel of Figure \ref{fig:ui}, labeled (A), displays $25\,600$ time series randomly drawn from the overall train and test data of the electricity dataset from the UCI repository\cite{electricityloaddiagrams20112014_321}, projected onto a two-dimensional PCA space. We choose this number of series to facilitate the visibility while capturing the true shape of the distributions. Here we have selected the test time series with label 1272 for illustration, highlighted in the PCA space by a green dot. In this panel, users can choose between the PCA-plot shown in Figure \ref{fig:ui} and the histogram seen in Figure \ref{fig:histogram}. Time series 1272 is displayed in the bottom panel, labeled (B) in Figure \ref{fig:ui}, in green, with a superposed pink line denoting the model forecast for the last 24 hours of series. Panel (D) in Figure \ref{fig:ui} shows the numeric feature values of the dataset and the selected time series. Together with panel (A) and (B), it allows users to visualize and explore datasets (\textbf{R1}), and do a qualitative assessment of forecasting performance at specific areas of the instance space (\textbf{R2}, \textbf{R3}).

The green line in the top right panel, (C) in Figure \ref{fig:ui}, shows the MASE score for the forecast, and the orange line and bands denote the average and 50\% confidence interval, respectively, for the score of the entire test set for comparison. Because the selected time series is located in the region of the PCA space well-covered by the training data, its prediction score in the top right panel lays mostly at a level comparable to that of the mean, except around horizon 20. By choosing the "Metric scores" tab in panel (D), users get exact numeric values for both the training and test dataset, and the selected time series, and users can select between multiple error metrics. Thus \texttt{EXPRTS} provides information that allow users to evaluate a forecasting model performance and robustness at multiple levels, from an aggregate over multiple instances and down to errors on individual horizons (\textbf{R2}, \textbf{R3}).

Panel (E) in Figure \ref{fig:ui} shows a set of transformations, which were introduced in Section \ref{sec:ts-features}. The three first transformation sliders correspond to changing $f$, $m$ and $h$ in Equation \ref{eq:t-transform}, while the forth slider corresponds to changing $k$ in Equation \ref{eq:s-transform}. The last two sliders gives users the option to add Gaussian noise to the time series, and the user can choose the fraction of points in the time series to be perturbed and the degree of the perturbation. The tab named "Local transformations" allows the user to add noise, shift the signal vertically, and change the seasonality strength as described before, but limiting these transformations to only a part of the time series selected by the user (we show examples of how these transformations can be used in the use case in section \ref{sec:use-case3}). Lastly, a tab named "general transformations", accessible by using the page-navigation icon in panel (E), permits users to vertically shift and rescale the whole time series by adding or multiplying the time series with a constant value. This value can be input by the user as a number or selected with a slider. Defining so many transformations enables users to probe a forecasting model's robustness to many different kinds input time series, uncovering the types of series a model can reliably forecast (\textbf{R2}).

\section{Use cases}\label{sec:use-cases}
To illustrate the usefulness of \texttt{EXPRTS}, we present three use cases. The use cases show how \texttt{EXPRTS} can help users explore individual instances in a dataset, asses forecasting performance on particular types of time series and generate new data exhibiting specific characteristics. In the last use-case, we showcase how these ideas can come together to improve the robustness of a forecasting model.

\subsection{Generating new data with specific characteristic}\label{sec:use-case1}

\begin{figure}
    \centering
    \subfloat[]{
        \includegraphics[width=0.8\linewidth]{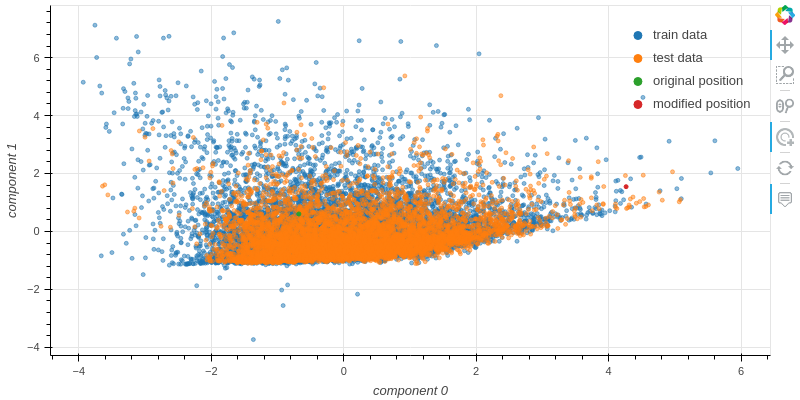}
        \label{fig:sanfran-a}
    }
    \\
    \subfloat[]{
        \includegraphics[width=0.8\linewidth]{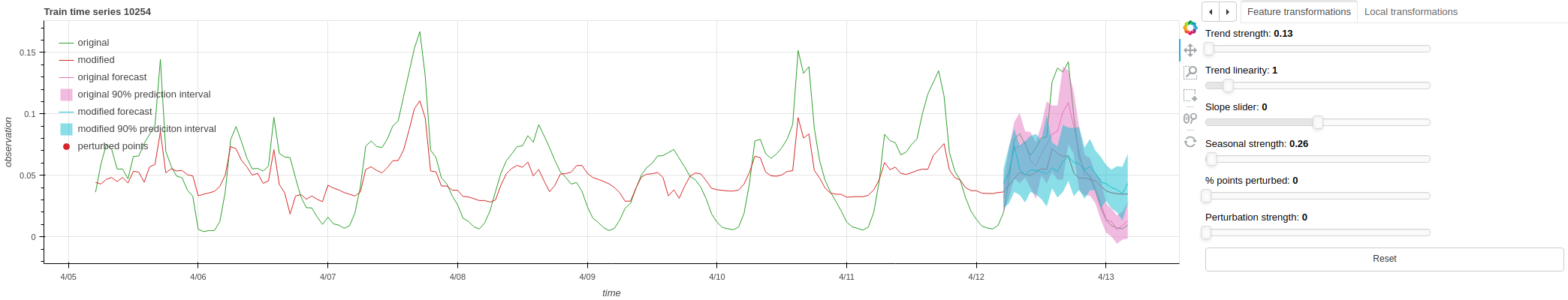}
        \label{fig:sanfran-b}
    }
    \caption{a) The instance space for the San Francisco Traffic dataset. The selected time series is moved from the green dot to the red in the instance space b) The original and transformed time series, together with the transformations applied to generate it.}
    \label{fig:sanfran}
\end{figure}

In the first use case we meet Alice, a machine learning engineer working for a company specializing in route assignment of trucks. To find the optimal route assignment, the company must forecast the traffic load accurately. However, the traffic loads for multiple roads is expected to change due to maintenance at a large intersection. The road work is expected to make the traffic patterns noisier as various roads open and close throughout the period. Alice's employer wants ensure that their traffic load forecast stays accurate when the maintenance work starts, and Alice is tasked with generating new training data for the model to ensure that is the case.

Alice decides to use \texttt{EXPRTS} to visualize the instance space of the dataset of traffic patterns\footnote{Here we use the dataset San Francisco Traffic, available through the Monash Time Series Archive\cite{monash_repository}, as an illustration.}. The instance space can be seen in Figure \ref{fig:sanfran-a}. By visual inspection, Alice finds that the series the more noisy series in the dataset are located to the far right in the instance space. Having identified the area of interesting time series, she uses \texttt{EXPRTS} to prototype the kinds of transformations she needs to apply to time series to generate series in the same region of the instance space. She finds that using the transformations to reduce both the trend strength and seasonal strength produces the desired effect, which is consistent with her wish of creating noisy time series. Figure \ref{fig:sanfran-b} shows an example of the types of time series she wants to generate. As a result, Alice is assured that she can generate data that captures the expected change in traffic patterns. After generating a new dataset, she retrains the forecasting model and remains confident that will continue to perform well in the future.

\subsection{Location dependent performance on the monthly subset of M4}\label{sec:use-case2}
\begin{figure}
    \centering
    \includegraphics[width=\linewidth]{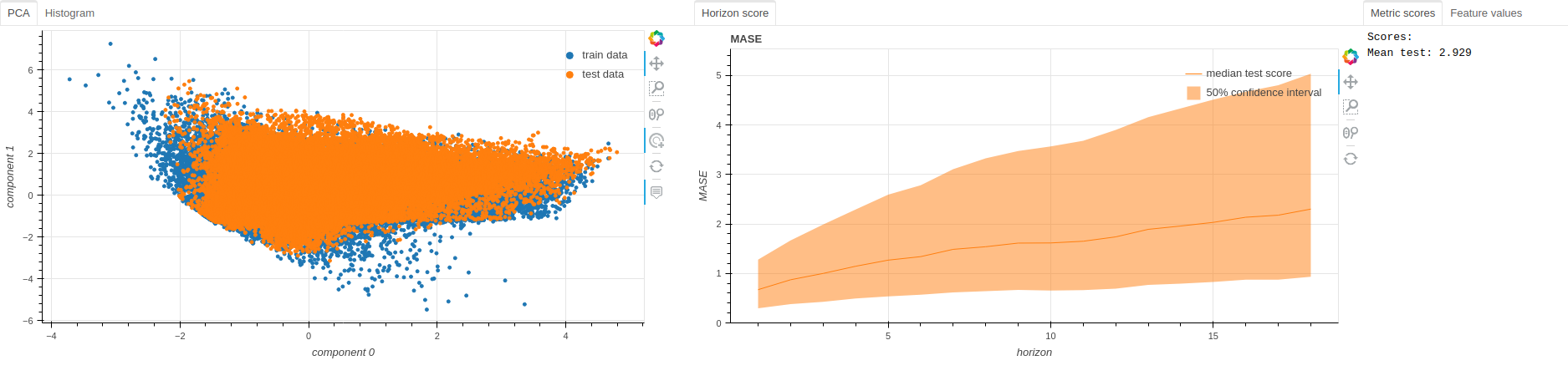}
    \caption{The instance space, error plot and error scores for the monthly subset of the M4 dataset.}
    \label{fig:m4}
\end{figure}

In this use case, Bob, a machine learning PhD student, is asked to investigate the performance of a model on the M4-monthly time series \cite{makridakis_m4_2020}. The dataset contains 48 000 different time series of various lengths and characteristics, making it possible for Bob to asses performance under several different types of conditions. At first, Bob looks at the instance space of the data, seen in Figure \ref{fig:m4}. By looking at the projection of the data in the PCA space, Bob notices that the test data occupying the region with the second PCA component (the vertical axis) larger than $4.5$ is barely covered by training data. Bob thinks that such a lack of coverage of training data could indicate that the forecasting model performance is worse in the region. A visual inspection using \texttt{EXPRTS} reveals that the series that Bob have identified all contain an upward, linear trend, and the model appears to struggle to predict such time series.

To investigate his hypothesis, he compares the average MASE value of the test data and the 17 time series in the aforementioned region. Figure \ref{fig:m4} shows that the average MASE value on the test data is $2.929$. In comparison, Bob calculates the average MASE value for the selected 17 time series and finds a value of $6.905$. Thus Bob's suspicion is confirmed; the model performance for the test data in the under-sampled region is significantly lower than the average for the whole test set, and \texttt{EXPRTS} helped Bob discover a subset of time series where the forecasting model is performing much worse than the average.

\subsection{Improving robustness of models for electricity consumption forecasting}\label{sec:use-case3}
As our last use-case, we do an empirical experiment where we show how \texttt{EXPRTS} can be used to improve the robustness of a forecasting model. We use the electricity dataset\cite{electricityloaddiagrams20112014_321}, containing the electricity consumption of 370 different consumers  and want to address the following question: can we, guided by the instance space and transformations defined in \texttt{EXPRTS}, generate data at a undersampled region in the instance space and use it to improve model performance to out-of-sample series in the same region? That is, to what extent do the simple transformations used by \texttt{EXPRTS} constitute a valid training sample? To do so, we want to find a subset of our test data that is situated in a region of the instance space not covered by the training data. Then, we will generate data to cover the subset of test data, retrain the model with the generated data, and lastly compare the results on the subset of test data.

\begin{figure}[t]
    \centering
    \includegraphics[width=\linewidth]{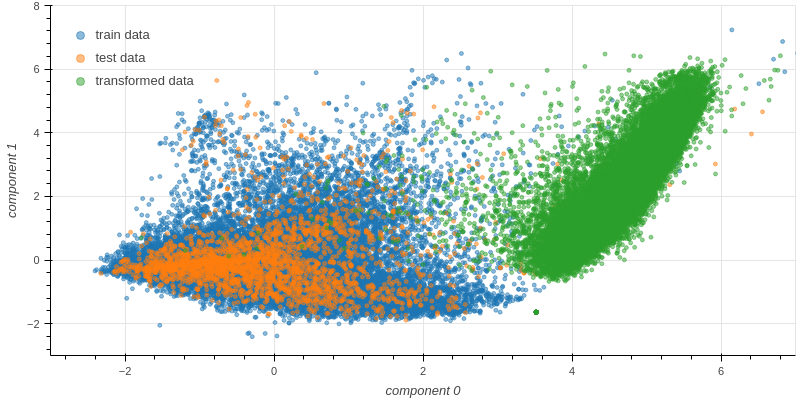}
    \caption{The instance space of electricity consumption data, after adding transformed data that contains sudden jumps in level.}
    \label{fig:transformed-data}
\end{figure}

As a first step, we identify a region in the instance space that contains test data samples not yet covered by training data. In Figure \ref{fig:ui} we see that the test data occupying the region with the first PCA component (the horizontal axis) greater than four is such a region. There are 13 time series in the selected region, and, by using \texttt{EXPRTS} to do a visual inspection, we identify that all of them exhibit a sudden jump while the "normal" time series do not. On the entire test dataset, our model has an average MASE value of $1.019$. However, on these 13 time series the average MASE is $34.168$, with the median being $6.485$, confirming that our model perform substantially worse in this region of the instance space. From now on, the 13 time series will be referred to as the subset of test data.

To generate new data similar to the subset of test data, we find a suitable transformation in \texttt{EXPRTS} and apply to every individual time series in the training data. Specifically, we use \texttt{EXPRTS} and select a time series in the center of the distribution in the instance space. By using the local transformations of \texttt{EXPRTS}, we increase the level of the latter half of the time series. Such a transformation moves the selected time series to the region of our subset of test data, suggesting that it is similar to our subset of test data. Finally, we transform every time series in our training data in a similar way, adding some noise in the transformations to induce variability (see Appendix \ref{sec:full_user_test_details} for the exact details of the experiment). The instance space containing the training, test, and generated data can be seen in Figure \ref{fig:transformed-data}, which shows that the generated data approximately covers the region of instance space we targeted.

With new data generated, we retrain our model using two different datasets: one dataset containing only generated data, and one dataset containing both the original training data and the generated data. The MASE of the different models are shown in Table \ref{tab:mase}. The table shows that using the generated data alone significantly improves the mean value of the MASE score compared to the original training set. This (unsurprising) result is driven by the fact that we have now trained a neural network to predict series similar to the ones in our subset of test data. Note that there are some outliers in our subset of test data which are not sufficiently covered by the data we have generated. These series drive the large mean values seen for in Table \ref{tab:mase}, but for the remaining time series, which are now better covered by the transformed data, the improvement is large. The effect is illustrated by a reduction of the median MASE value by a factor of almost three; because the median is less sensitive to outliers than the mean, the former reflects more accurately the overall improvement for the majority of the 13 time series.

\begin{table}[!t]
    \caption{MASE forecasting performance for the 13 time series in the undersampled region with the first PCA component $>4$, by using the original, transformed, and both original and transformed training datasets.}
    \label{tab:mase}
	\centering
    \begin{tabular}{
    	*{4}{c}
	}
    \hline
        Train data & Mean &   Median &   Std   \\
    \hline
        Original & 34.168 & 5.071 & 113.732 \\
        Transformed & 31.394  & 2.244 &  114.762 \\
        Orig + Trans & 31.982  & 2.815   & 114.265 \\
    \hline
    \end{tabular}%
\end{table}

For comparison, the bottom row of Table \ref{tab:mase} quotes the MASE values from using both the original and transformed, training sets to train the model. Despite the fact that the amount of training data is now doubled, the results are slightly worse than the case with the transformed data alone. This highlights the higher importance of the location of the data in the feature space (and in turn of the similarity between training and test data) compared to the amount of data alone. Including a large amount of original data far from the 13 test time series (blue points) forces the model needs to learn to forecast a more diverse set of time series, at the expense of precision in the smaller region of interest (see \cite{semenoglou_data_2023} for a discussion the importance of the location of training data in the instance space).

\section{User Tests}\label{sec:user-tests} 
To evaluate \texttt{EXPRTS}, we conduct user tests with professionals specializing in machine learning engineering and time series analysis. Each test lasted one hour, and the users were asked to complete a series of tasks, share their feedback, and respond to a set of predefined questions. The tests aims to evaluate if \texttt{EXPRTS}: 1) provides users with insights into the distribution of the dataset, 2) provides users with insights into the predictions of the forecasting model, and 3) if the use of time series transformations provide users with insight into the robustness of the forecasting model. In this section, we describe and evaluate the test design, and present and discuss the test results.

\begin{table}[tb]
    \caption{Summary of user test tasks}
    \label{tab:user_test_tasks}
    \centering
    \begin{tabular}{p{2.0cm}p{6.0cm}}
    \hline
      Topic & Description \\ 
    \hline
    PCA Space & Inspect time series with increasing value of component 0 (x-axis) in the PCA space, then do the same for component 1 (y-axis). Which features do the components describe? What features is most prominent in the dataset? \\
    Predictions & By interacting with the PCA plot, identify areas where the model struggles with its predictions. Why do you think the model struggles in these areas of the PCA plot? \\ 
    Transformations & Select the time series with index 113, then transform the 4 feature values of the time series. How could these transformations be used to gain insights into potential future scenarios? How could you use this to evaluate the model's performance on different inputs? Do you understand what the transformations mean, and do they behave as you expect? Transform time series 113 to the section of the PCA space where component 0 is between 4 and 6 and component 1 is between 2 and 4.\\ 
    \hline
    \end{tabular}
\end{table}
\subsection{Test Design}
The test is designed as a qualitative, semi-structured interview, where users provide feedback while completing predefined tasks and answer a set of predetermined questions.  A total of 10 users participated in the user test. The users are professional machine learning engineers or time series analysts, with no previous exposure to the \texttt{EXPRTS} tool. The tasks are designed to resemble realistic use cases of the tool, and are summarized in Table \ref{tab:user_test_tasks}. The goal of each task can be summarized as follows: the \textbf{PCA Space} task examines the insights provided by visualizing the time series dataset in a 2D PCA space; the \textbf{Predictions} task investigates the relationship between time series features and model prediction errors; and the \textbf{Transformations} task evaluates the value added by applying feature transformations to create new time series.


The user test consist of two main phases: \textit{introduction} and \textit{test}. In the \textit{introduction} phase, the user is first introduced to the time series features and then guided through the different parts of the user interface of \texttt{EXPRTS}. The \textit{test} phase consist of the three tasks, which are summarized in Table \ref{tab:user_test_tasks}. During each task, the user provides feedback as the task is being solved, and in the end interview with a set of predefined questions. The statements are listed in Table \ref{table:question_mapping}, and the possible answers to each statement are: \textit{Agree}, \textit{Partially Agree}, \textit{Partially Disagree}, and \textit{Disagree}. For a full overview of the different tasks and questions in the user test, along with configuration details of \texttt{EXPRTS} we refer the reader to Appendix \ref{sec:full_user_test_details}.  

\subsection{Evaluation of Test Design}
The test design is based on the ``Minimum Necessary Rigor'' framework proposed by Klein et al. \cite{klein_minimum_2023}, which prioritizes obtaining valuable insights efficiently from user tests of human-AI systems. The goal of this user test is to explore how users perceive \texttt{EXPRTS} and to determine whether, and how, it provides them with valuable insights. To this end, we use a semi-structured interview format for the user test to gather the necessary feedback. Other studies presenting comparable AI systems and frameworks \cite{yeh_attentionviz_2024, cheng_dece_2021, spinner_explainer_2020} employ a strategy similar to that of this user test, collecting qualitative data through user feedback.

\begin{table}[tb]
  \centering  
  \caption{\label{table:question_mapping} Mapping of statement codes to full statements.}
\begin{tabular}{p{0.5cm}p{7.5cm}} 
    \hline
    Code & Full statement \\ 
    \hline
    S1 & By interacting with the dataset through the PCA and the time series plot, it is easy to understand how the time series features are represented in the dataset. \\ 
    S2 & The information provided by \texttt{EXPRTS} enables me to easily evaluate how the model performance depends on time series features. \\
    S3 & Understanding how model performance depends on time series features is useful when developing forecasting models. \\
    S4 & I am able to anticipate how changing the \textbf{trend strength} will transform a time series. \\
    S5 & I am able to anticipate how changing the \textbf{trend linearity} will transform a time series. \\
    S6 & I am able to anticipate how changing the \textbf{trend slope} will transform a time series. \\
    S7 & I am able to anticipate how changing the \textbf{seasonal strength} will transform a time series. \\
    \hline
\end{tabular}
\end{table}

\subsection{Results \& Discussion}
\begin{figure}[ht]
  \centering
  \includegraphics[width=0.48\textwidth]{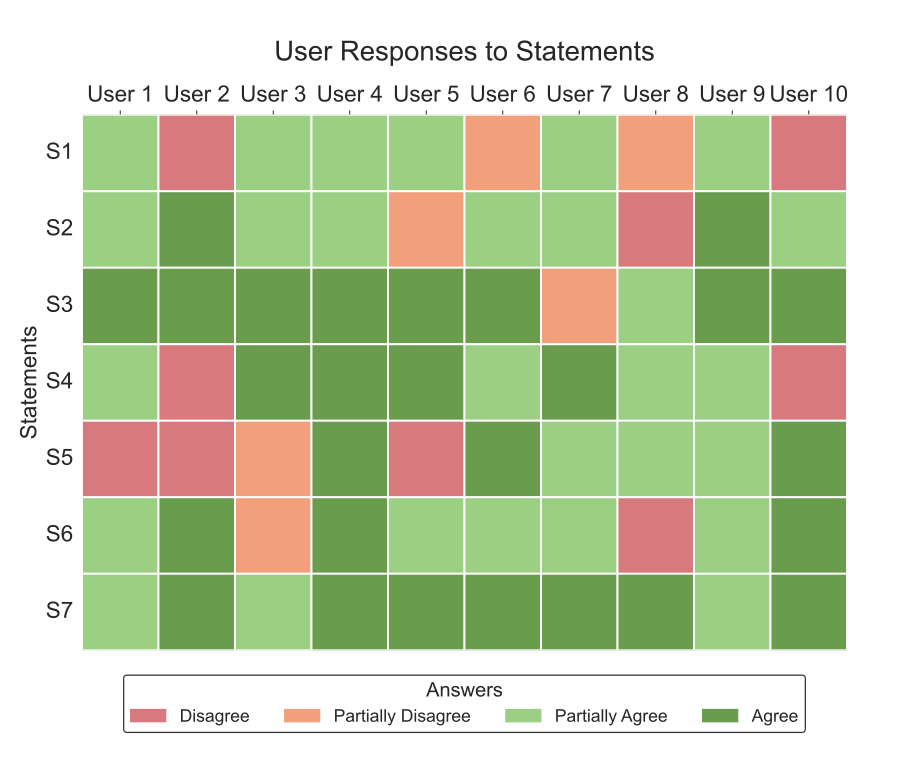}
  \caption[Users' answers to statements given during user test]{\label{fig:discreteresults} The answers of statements ranging from disagree-agree. See Table \ref{table:question_mapping} for full statement.}
\end{figure}

In this section we present and discuss the results of the user test task, with the users being referred to as U1 - U10. Figure \ref{fig:discreteresults} presents the answers given to each statement in Table \ref{table:question_mapping}. 
 
\subsubsection{PCA Space}
The concept of representing the time series in the dataset in terms of their features is in general easy for the user to grasp, but some users require more time for this than others. The users identify that the dataset contains time series with a varying degree of seasonal strength and trend by manually inspecting the characteristics of a number of different time series in the PCA space. Most users refer to the trend as one concept, and have difficulties with separating it into trend strength, trend linearity and trend slope. As shown in the row marked S1 in Figure \ref{fig:discreteresults}, 60\% of users partially agree that the PCA space visualization simplifies identifying the dataset's key characteristics. No user fully agree with the statement. When asked why, users say it is hard to connect the original time series features to the PCA components.


\begin{figure*}[tbp]
    \centering
    \subfloat[]{
        \includegraphics[width=0.48\textwidth]{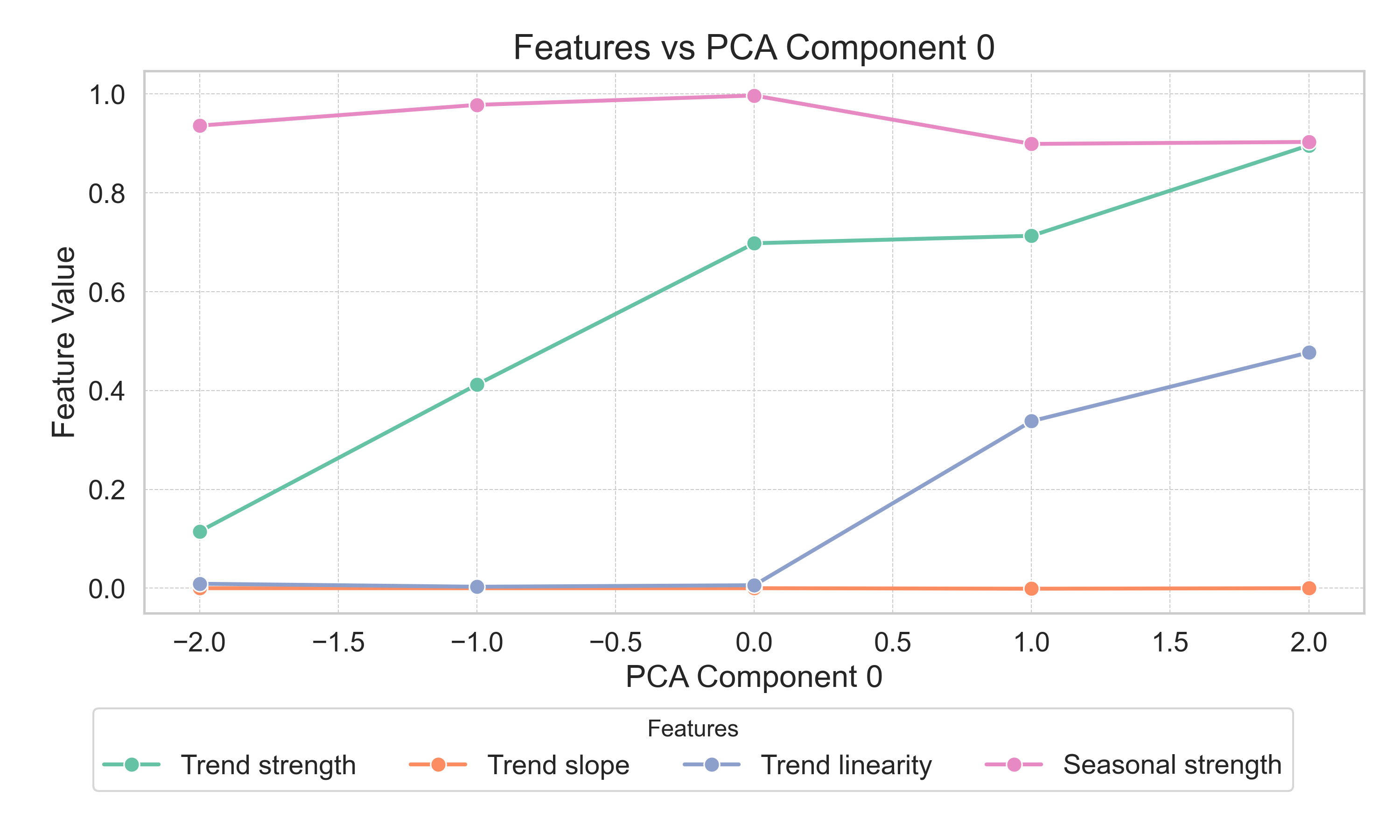}
        \label{fig:pca0_changes}
        
    }
    \hfill
    \subfloat[]{
        \includegraphics[width=0.48\textwidth]{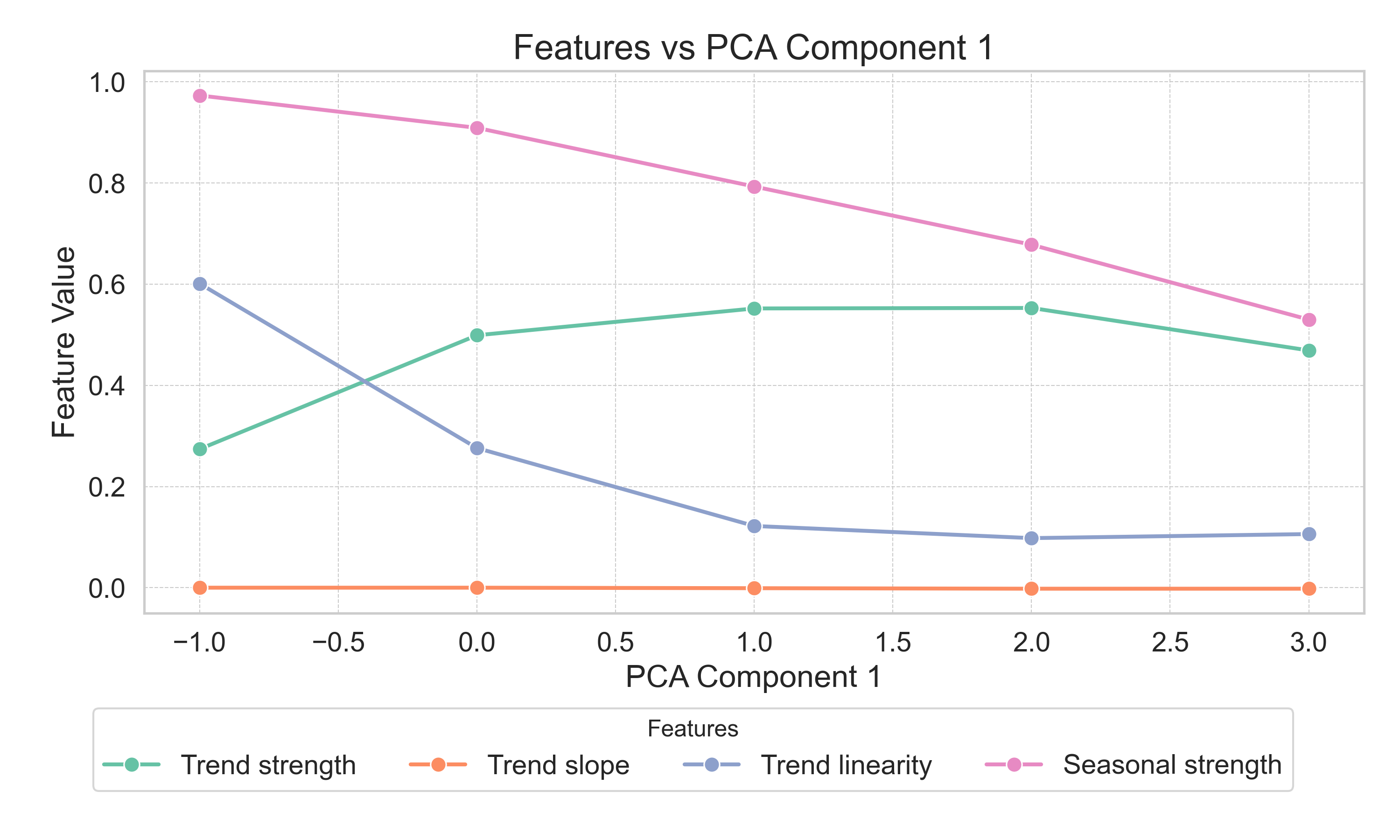}
        \label{fig:pca1_changes}
    }
    
   \caption[Feature values plotted against PCA components]{The two plots illustrate a typical scenario for users while they explore the relationship between the PCA componenets and the time series features. Time series were selected with a varying PCA component values. (a) illustrates feature values of time series when keeping PCA component 1 constant equals to 0 and increasing PCA component 0. (b) illustrates feature values of time series when keeping PCA component 0 constant equals to 0 and increasing PCA component 1. The selection of time series and extraction of time series feature values are done in the \texttt{EXPRTS} user interface to resemble what was done during the user test. The changes of trend slope and trend linearity in (a) is not linear, and quite hard for users to interpret, while the changes to seasonal strength in (b) makes it easier to extract this feature from the PCA component.}
    \label{fig:example_pca_explore}
\end{figure*}

To better understand the difficulties of identifying the relationship between the PCA components and the time series features we investigate how each time series feature change with the PCA components. When attempting to understand the component-feature relationship, users typically looked at time series with a fixed value for one component, and increasing values of the other component. Figure \ref{fig:example_pca_explore} illustrates the situation where one of the PCA components is changed while the other is kept fixed at 0. Observe that in Figure \ref{fig:pca0_changes}, as PCA component 0 increases, the trend strength becomes larger, but when the value of component 0 reaches a value larger than 0, the trend linearity starts to increase as well. We conjecture that such irregular changes in feature values makes it harder for users to understand and relate the features to areas in the instance space. A similar investigation of PCA component 1, shown in Figure \ref{fig:pca1_changes}, reveals that as PCA component 1 increases, seasonal strength and trend linearity decreases. Users typically only make the connection between component 1 and seasonal strength, with no mention of trend linearity. We suspect this is because the visual impact of varying seasonal strength in the time series is more pronounced compared to that of trend linearity.

\subsubsection{Predictions}
As shown by S2 in Figure \ref{fig:discreteresults}, 80\% of users agree to some degree that \texttt{EXPRTS} helps highlighting the relationship between the performance of the forecasting model and the time series features. When asked why, users note that they are able to identify that the performance is worse for higher values of trend related features, and better for time series with high seasonal strength. Most users also note that the performance is worse for outlier time series. However, note that users only \textit{partially agree} with S2 while many of the same users \textit{agree} with S3. We interpret the discrepancy between S2 and S3 as users recognizing the value of the proposed solution but not being entirely satisfied with how \texttt{EXPRTS} delivers these insights. The user who disagrees with S3 reasons that they would only find such information necessary when creating a universal forecasting model that performs well across time series with varying characteristics.

\subsubsection{Transformations}
In the final task of the user test, the users experiment with the transformations for \textit{trend strength}, \textit{trend linearity}, \textit{trend slope} and \textit{seasonal strength}. The users transform one feature at a time, while paying attention to the resulting changes in the time series and its forecasts. As shown by S4-S7 in Fig. \ref{fig:discreteresults}, the seasonal strength transformation (S7) behaves as expected. Meanwhile, the transformations of the trend related features are harder to anticipate and understand. Trend linearity is the most difficult (S5), while trend slope (S6) and trend strength (S4) are easier to understand. We would argue that the seasonal strength is easier to understand because it is the only feature related to the seasonality of a time series. While the concept of a time series' trend is well-known, the distinction between trend strength, linearity, and slope is new to most users, and the interdependence among them decreases the user's understanding. For instance, U5, U6, and U8 mentioned that this interdependency made the trend related feature transformations harder to understand. A clear example of this interdependency is observed when users make significant changes to the trend linearity of a time series while noticing only minor changes in the resulting time series. However, after applying transformations to the other trend features, the impact of the trend linearity transformation becomes more evident.

In the \textit{Transformation} task, users are asked to transform a time series so that the transformed version falls within a predefined area of the PCA space. To solve this task, most users chose a strategy of performing random transformations, observing the effects in the PCA space, and iteratively updating the transformations until the time series reach the desired area. Users choosing such a strategy suggests that they struggle with making targeted transformations. In total, \texttt{EXPRTS} provides functionality to transform one time series into another, but it does not effectively support users in directing these transformations toward a time series with a defined set of properties.

Compared to the \textit{Predictions} tasks, users find that probing the model to understand the relationship between model performance and time series features is easier when using the feature transformations. The transformations help most users to distinguish the trend features, which allow for a more detailed investigation of the performance-feature relationship. In the \textit{Predictions} task, users generally observe that model performance worsens as the trend becomes more prominent. After using the transformations, most users are able to identify that model performance worsens with increasing trend strength and slope, as well as decreasing trend linearity. This suggests that the transformations assist users in gaining a better understanding of each feature, the instance space and its relationship to model performance. We attribute this to the fact that in the \textit{Transformation} task users focused on one time series at a time, and by applying transformations, they immediately observe the effects on both the time series and the model's forecast. This enables users to quickly test their hypotheses about the features and their impact on the forecast. In this way, the feature transformations helped address some of the issues users encountered in the \textit{Predictions} tasks.

\section{Discussion, Limitations and Future Work}

\texttt{EXPRTS} is a flexible tool that can be used with to explore a wide variety of time series datasets and probe the robustness of forecasting models. Overall, the feedback on \texttt{EXPRTS} was positive; a large portion of the users found our transformations useful and said that \texttt{EXPRTS} allowed them to evaluate a time series forecasting model on time series with different features.

One of the main challenges identified was that users struggle to understand the relationship between the two PCA components and the four time series features. This suggests that improving the visualization or explanation of the PCA space would be helpful to users. The issues could be addressed by using a different dimensionality reduction technique like t-SNE\cite{maaten_visualizing_2008}, or by choosing a different set of features. On one hand, the first approach as been used in a similar context by \cite{kang_gratis_2020}, but since t-SNE does not preserve global structure between points we find it unsuitable for our use-case. Using a different set of features is easy to accommodate in our framework, but comes with the downside of the transformations loosing some of their immediate semantic meaning. However, as user-tests show, the users found the trend related transformations difficult to understand. While we argue this difficulty is partially explained by the users being unfamiliar with the tool, introducing transformations and features that are independent to a larger degree appears to be a promising avenue for improving both our framework and \texttt{EXPRTS}.

The feedback from the user tests indicate that transforming time series features helps users gain a better understanding of how those features relate to the forecasting model's performance. A key limitation of \texttt{EXPRTS} that emerged from the test was its inability to provide an easy way for users to transform a time series into one with a desired set of properties. This represents a usability gap that could be addressed to enhance the tool's flexibility and ease of use. Solving this issue is not trivial; changing any component of a time series, either the trend, seasonality or residual, has the potential to affect the other components. Furthermore, our visualization first compresses a time series into just 4 features, of which only the first 2 PCA components are visualized. Going backward from a specific point in the instance space to a full-length time series could be done using an time series autoencoder as in \cite{ali_timecluster_2019, rodriguez-fernandez_deepvats_2023}, but such an approach would introduce a lot of complexity.

Finally, we've only considered univariate time series datasets and forecasting models. This is a major limitation that stops us from applying our framework and \texttt{EXPRTS} to a large number of interesting, but multivariate forecasting problems. Extending our work from univariate to multivariate time series would require us to rethink the user-interface of \texttt{EXPRTS}, and include new visualization techniques to visualize multiple time series at the same time. Furthermore, a framework for transforming multivariate time series should ensure that the relationships between the individual series remain consistent and realistic, making the problem much harder. Hence, we view the task of extending our work to accommodate multivariate time series as important and interesting future work.

\section{Conclusion}\label{sec:conclusion}
We have presented a framework for generating semantically meaningful time series data and the tool \texttt{EXPRTS}. Our framework can generate a wide variety of different types of time series quickly, and does so through simple transformations that are easy to interpret for users. \texttt{EXPRTS} allow users to explore time series datasets and probe the robustness of neural network models used in time series forecasting. Together, they enable users to explore time series datasets, generate new time series and evaluate model performance in hypothetical scenarios. Assessing the robustness of these models under hypothetical scenarios is important because thee model might perform worse than expected when the data distribution changes. In real-world scenarios this translates into poor forecasts, and it is difficult to make informed decisions to ameliorate the consequences ahead of time. \texttt{EXPRTS} allows the user to explore these hypothetical scenarios by creating new time series and evaluating model performance.

We demonstrated our framework for generating data through an empirical experiment, showing how we can generate out-of-distribution data with specific characteristics to increase the robustness of a forecasting model in key situations. Afterwards, we evaluated \texttt{EXPRTS} through user-testing. The tests show that, overall, \texttt{EXPRTS} enables its users to gain insight into the aspects that drive the performance of models in time series forecasting tasks. The information provided by \texttt{EXPRTS} can be used to explore alternate scenarios in an interpretable way, as well as to improve the performance of forecasting models in certain regions of the feature space. These qualities allow the user to explore and anticipate the impact of potential events and, therefore, make informed decisions and take appropriate actions in both, academic and business settings.

\bibliographystyle{abbrv}
\bibliography{references}

\appendix
\section{Appendix}
\subsection{Use Case Experiment Details}
\label{sec:use_case_experiment_details}
\begin{figure}
    \centering
    \includegraphics[width=\linewidth]{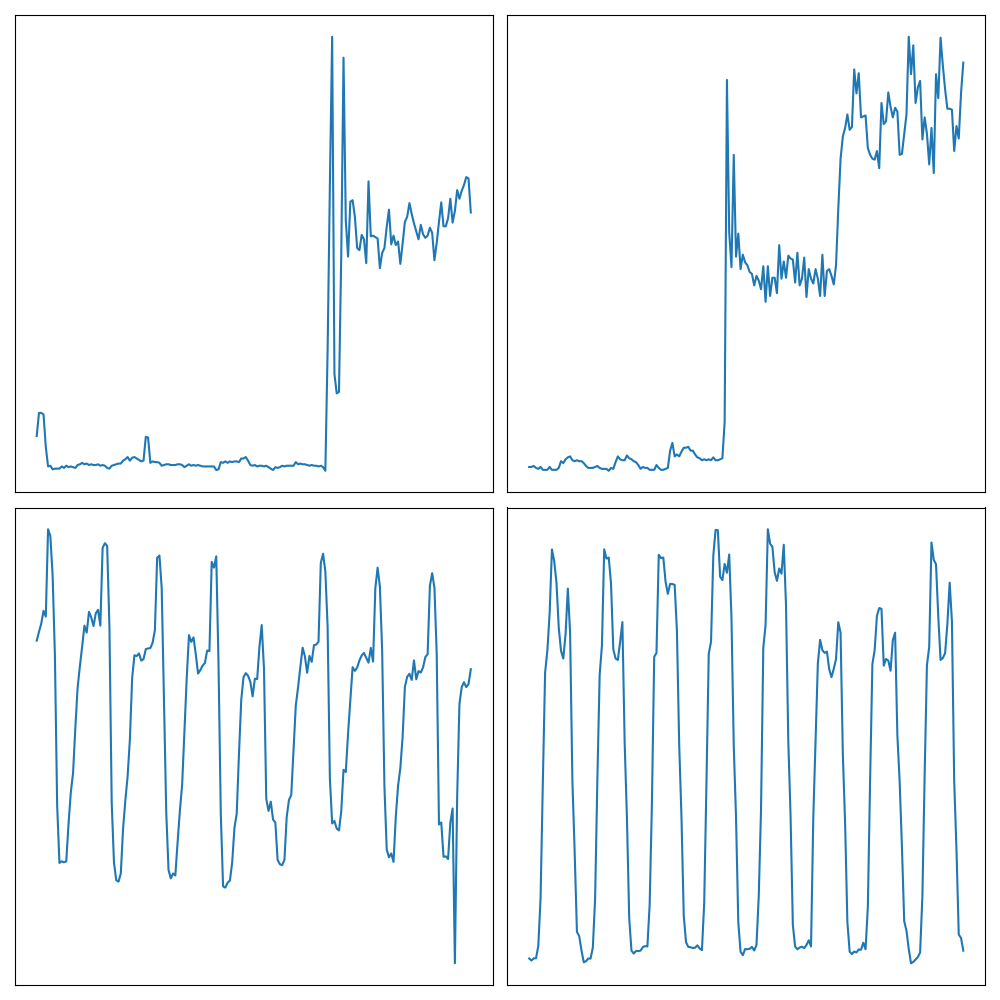}
    \caption{Four different time series from the electricity dataset. The top row depicts time series from the undersampled region of the instance space. The bottom row shows time series from the center of the data distribution in the instance space.}
    \label{fig:jump_examples}
\end{figure}

In order to allow for variability in the new (transformed) data, we apply the following transformations to each training time series: we transform the part of the time series with time above an integer value uniformly drawn between 72 and 144. Because these two values refer to hours (which is the unit resolution of the time series) ranging from 0 to 168, this corresponds to the part of the time series that starts between the beginning of the fourth day and the end of the sixth day, and extends to the end of the time series. This choice is motivated by the observation that the upward jump in the upper time series of Figure \ref{fig:jump_examples} usually occurs after a few days. A more accurate assessment of the exact changes may be preferred in reality but we chose this simple approach here for illustration. Next, the boost of the average signal applied to the selected region of each time series is a multiplicative factor drawn from a uniform distribution covering the range $2-5$. We have explored the impact of various multiplicative factor values on a few time series by visualizing the effect  with \texttt{EXPRTS}, and have found that values above $6-7$ do not produce further significant differences; in these cases the effect saturates and, if applied to all the training set, most of the time series appear to cluster around a value of the PCA component 0 of $5.5 - 6$, leaving regions with lower component values undersampled.

In the experiment we use a dense neural network, and the parameters are shown in Table \ref{tab:model_training_args}.

\subsection{User Test Details}
\label{sec:full_user_test_details}
This appendix covers the configuration of \texttt{EXPRTS} and the set of instructions and questions used by the interviewers during the user tests.

\subsubsection{Configuration of \texttt{EXPRTS} during user tests}
\label{sec:usertestconfig}

For reproduction purposes, we present the configuration of \texttt{EXPRTS} during the user tests. Regarding the model, a feed forward model, with parameters given in Table \ref{tab:model_training_args} is used. For the time series data, the \textit{electricity nips} dataset which is provided by GluonTS \cite{gluonts_jmlr} is used.
\begin{table}
    \centering
    \caption{Model Architecture and Training Arguments}
    \begin{tabular}{ll}
        \hline
        \textbf{Model Architecture}                & \textbf{Details} \\ \hline
        Input layer                                & 168 neurons       \\
        Hidden layer 1                             & 100 neurons       \\
        Hidden layer 2                             & 100 neurons       \\
        Output layer                               & 24 neurons        \\ \hline
        \textbf{Training Arguments}               & \textbf{Details} \\ \hline
        Random seed                                & 0                 \\
        Seasonal Periodicity (sp)                  & 24                \\
        Batch size                                 & 512               \\
        Epochs                                     & 100               \\
        
      Number of batches per epoch                & 50                \\
        Number of batches to write                 & 0                 \\
        Number of validation windows               & 7                 \\
        Early stopping patience                    & 10                \\
        Allow padded sampling                      & False             \\
        Scaler used                                & Standard          \\ \hline
    \end{tabular}
    \label{tab:model_training_args}
\end{table}

\subsubsection*{Introduction}
A common issue for machine learning models applied to time series forecasting is the temporal evolution of the dataset distribution. When facing such situations, it would be valuable to reliably anticipate the impact of the distribution changes ahead of time.\\
This tool allows users to understand how the performance of a time series forecasting model depends on the features of the time series in the dataset. Additionally, new time series can be generated by altering the features of existing time series.

Representing time series in terms of their features is an essential concept in the context of this tool. The features are based on seasonal-trend decomposition of time series and can be described as follows: 
\begin{itemize}
    \item Trend strength: Describes to what degree the the trend component contributes to the time series compared to the residual component.
    \item Trend linearity: Describes to what degree the trend component resembles a linear function.
    \item Trend slope: Describes the slope of the trend component.  
    \item Seasonal strength: Describes to what degree the seasonality component contributes to the time series.\\
\end{itemize}
Throughout this test, these characteristics are referred to as features.
\\\\
The main components of the user interface are described as followed.
\begin{itemize}
    \item PCA plot: The upper left plot shows each time series as a distinct point. Each time series is expressed in terms of their feature values, and then processed with PCA to allow visualization in a 2D space.
    \item Time series plot: The bottom plot displays the selected time series along with its prediction. Additionally, it shows the transformed time series if any transformations has been made.
    \item Horizon score plot: The upper right plot illustrates the error distribution of the test data predictions. Additionally, it shows the error of a particular chosen or transformed time series prediction when applicable.
    \item Numeric info tabs: On the right-hand side of the horizon score plot there are two tabs displaying numeric information about errors and features. Metric scores displays the mean error of the test set, the selected time series, and the transformed time series. Feature values displays info about the feature values of the train and test set, as well as the exact feature values of the selected and transformed time series.
    \item Transformations tabs: On the right-hand side of the time series plot there are several tabs related to time series transformations. 
    General selections allows the user to select a specific time series by its ID. 
    General transformations allows the user to modify the selected time series by an additive or multiplicative constant, or both. 
    Feature transformations allows the user to modify the time series by altering the four time series features presented earlier. Additionally, the user may add random perturbations to the time series.
\end{itemize}

User tests are performed using the electricity data et and the feedforward model. 

Disclaimers:
\begin{itemize}
    \item In the following tasks and related questions, we are not seeking the ``correct answer", we are interested in your thoughts on the matter.
    \item The test runners have not been involved in the development of the tool and should be viewed as external testers. 
    \item Answer honestly. Whether or not the feedback is positive does not affect our academic results, and don't feel bad about giving negative feedback. All feedback is valuable.
\end{itemize}

Try to think aloud while working through the tasks.

\subsubsection*{PCA Space}
This task will not go over altering the time series in any way, only looking at how the different time series are placed in the instance space, and how moving along the two axes alters the time series. These are the tasks to perform:

\begin{enumerate}
    \item Inspect time series with increasing component 0 values, and then inspect time series with increasing component 1 values. How does changes in the component value affect the time series in terms of the time series features?
    \item Inspect the outlier points. How can these be characterized in terms of the time series features?
\end{enumerate}

After these tasks have been completed, answer these questions:

\begin{itemize}
    \item How does the features of time series change as component 0 is increased?
    \item How does the features of time series change as component 1 is increased?
    \item How would you characterize this dataset in general in terms of time series features?
    \item How would you characterize the outliers in terms of time series features?
\end{itemize}

Answer the following statement with either \textit{disagree, partially disagree, partially agree or agree}, please explain your answer:

\begin{itemize}
    \item By interacting with the dataset through the PCA and the time series plot, it is easy to understand how the time series features are represented in the dataset.
\end{itemize}

\subsubsection*{Predictions}
The goal of this task is to inspect the predictive performance on time series across different areas of the PCA space. These are the tasks to perform:

\begin{enumerate}
    \item Interact with the PCA plot. Try to identify areas where the model struggles with it's predictions.
    \item Try to identify how the quality of the prediction varies with respect to the time series features.
\end{enumerate}

After theses tasks are completed, answer the following questions:

\begin{itemize}
    \item In which areas in particular does the model struggle with its predictions?
    \item Why do you think the model struggles in these particular ares?
    \item How would you describe the performance of this model in terms of time series features?
\end{itemize}

Answer the following statement with either \textit{disagree, partially disagree, partially agree or agree}, please explain your answer:

\begin{itemize}
    \item The information provided by the tool enables me to easily evaluate how the model performance depends on time series features.
    \item Understanding how model performance depends on time series features is useful when developing forecasting models. 
\end{itemize}

\subsubsection*{Transformations}
In this task, the user will try to interact with the time series transformation feature of the tool. It is designed to highlight potential difficulties with the tool's user experience and how intuitive the transformation feature is for new users. These are the tasks to perform:

\begin{enumerate}
    \item Select the time series with index $113$ and locate it in the PCA space.
    \item For each transformation, describe: how it changes the time series; how it moves the time series in the PCA space; and how it affects the predictive performance.
    \item Transform the time series into a new time series placed in the region with a component 0 value between 4 and 6, and a component 1 value between 2 and 4.  
\end{enumerate}

After completing the tasks, answer these questions:

\begin{itemize}
    \item How would you use the transformations to gain insight into how the model performance depends on time series features?
    \item How would you use the transformations to improve the model performance on certain types of time series features?
\end{itemize}

Answer the following statements with either \textit{disagree, partially disagree, partially agree or agree}, please explain your answer:

\begin{itemize}
    \item I am able to anticipate how changing the \textbf{trend strength} will transform a time series.
    \item I am able to anticipate how changing the \textbf{trend linearity} will transform a time series.
    \item I am able to anticipate how changing the \textbf{trend slope} will transform a time series.
    \item I am able to anticipate how changing the \textbf{seasonal strength} will transform a time series.
\end{itemize}

\subsubsection*{Final Questions}
After completing every task, please answer these general questions regarding the tool:
\begin{itemize}
    \item What are your thoughts in general on the value of expressing time series in terms of the four features?
    \item After using the tool do you have any general feedback for improvements of the tool?
\end{itemize}

\end{document}